# 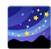 *emojiSpace*: Spatial Representation of Emojis

Moeen Mostafavi, Mahsa Pahlavikhah Varnosfaderani,
Fateme Nikseresht, Seyed Ahmad Mansouri
University of Virginia
{moeen, mp3wp, fn5an, sm2xv @ virginia.edu}

## Abstract

In the absence of nonverbal cues during messaging communication, users express part of their emotions using emojis. Thus, having emojis in the vocabulary of text messaging language models can significantly improve many natural language processing (NLP) applications such as online communication analysis. On the other hand, word embedding models are usually trained on a very large corpus of text such as Wikipedia or Google News datasets that include very few samples with emojis.

In this study, we create *emojiSpace*, which is a combined word-emoji embedding using the *word2vec* model from the Genism library in Python. We trained *emojiSpace* on a corpus of more than 4 billion tweets and evaluated it by implementing sentiment analysis on a Twitter dataset containing more than 67 million tweets as an extrinsic task. For this task, we compared the performance of two different classifiers of random forest (RF) and linear support vector machine (SVM). In the evaluation task, we compared *emojiSpace* performance with two other pre-trained embeddings and demonstrated that *emojiSpace* outperforms both.

## 1 Introduction

Recent research shows that expressing emotion can help people to improve their mental health [1]. Mehrabian et al. illustrated that 93% of the meaning related to emotion is transferred using non-verbal cues such as body language and vocal inflections [2]. However, these two are not part of the messaging communication mechanism. Emojis are a set of minor pictorial glyphs that help users express their emotions on messaging platforms. Emojis were first used in the late 90s and officially adopted into Unicode in 2010 [3]. Nowadays, people use emojis to convey their feeling while using messaging platforms.

Utilizing emojis in messaging has increased rapidly over the past few years. This increased usage led Oxford Dictionary to name 2015 the year of emojis [4]. One of the primary purposes of using emojis is to convey emotion during online communications. Users may even use more than one emoji in a single post to express their emotions and emphasize a certain point. A study by Riordan [5] showed that people would disambiguate the text using emojis. For example, if you use the text "Got a shot", adding an emoji of a ball, a glass of wine, or a syringe will help the reader to understand the text. The use of emojis in communicating with chatbots can improve the impression of the conversation for the customers [6]. Recent research shows how emojis can represent modifiers to express emotions during a conversation with chatbots [7; 8].

Online messaging has become the dominant form of communication recently [9]. As a result, there is a rapidly increasing interest in natural language processing (NLP) community to work on social media data. However, among several pre-trained word embeddings, just a few have an emoji embedding. For example, although *word2vec* was developed by Google in 2013 [10] and has been frequently used in literature, it includes some smileys and there is no emoji in its vocabulary. Similarly, *Glove* [11], another well-known word embedding model, was developed before emojis were frequently used. These two models were initially developed based on the data that includes very few emojis. Despite the users' interest in using emojis, many current language processing projects in the social media domain still utilize pre-trained word embeddings that do not include emojis in their vocabulary. In this project, we introduce emojiSpace, a word-embedding trained on billions of tweets and includes emoji in its vocabulary. We show how this embedding outperforms similar types of word+emoji embeddings.

## 2 Related Work

A few emoji embeddings are available in the literature, which are based on two main approaches: 1) utilizing social media data and finding the word embedding for all the words, including emojis [12; 13; 14]. 2) extracting labels that describe the emojis, and then the processed embedding of those labels represent the corresponding emoji embeddings [15]. The main limitation of the first approach is the limited number of tweets including emojis in the training set. For instance, Barbieri et al. used 100 M tweets, but only 700 of them included emojis [13]. On the other hand, since their dataset was significantly smaller than the ones used for training *word2vec* or *Glove*, the resulting embedding is not as general as *word2vec* or *Glove*. To overcome these deficiencies, researchers apply the second approach to extract emoji embedding.

One of the most frequently used emoji embeddings using the second approach is *emoji2vec* [15], which maps emojis into a 300-dimensional space that can be used along with Google News *word2vec* embedding. In *emoji2vec*, emojis' name and their descriptions were extracted from the Unicode emoji list. The authors defined vector representation of emoji names and descriptions, $v_j$, as the sum of embeddings for all individual word vectors in the *word2vec* space,

$$v_j = \sum_{k=1}^{N} w_k \qquad (1)$$

where $w_k$ is the embedding vector of $k^{th}$ description word if the word is in the *word2vec* vocabulary and 0 otherwise. Then the vector representation of emojis, $x_j$, is estimated based on maximizing the match between $x_j$ and $v_j$. For the training process they used the corresponding description vector of the emoji, $v_j$, as positive sample and a random sample description from other emojis, $v_i$ ($i \neq j$) as a negative sample [15]. In other word, they minimized the following loss function,

$$L(i, j, y_{ij}) = -y_{ij} \log\left(\sigma(x_i^T v_j)\right) - (1-y_{ij}) \log\left(\sigma(x_i^T v_j)\right) \qquad (2)$$

where $\sigma(x_i^T v_j)$ represent the sigmoid of the dot product between the two vectors and $y_{ij}$ is 1 when description $j$ matches emoji $i$ and 0 otherwise.

## 3 Methodology

In this project, we use the gensim[1] *word2vec* on a corpus of more than 4 billion English tweets to train the model. This large corpus of texts includes both words, emojis, and informal words. We used an extrinsic method (sentiment analysis) for the evaluation part and compared the performance of *emojiSpace* with two other embeddings using two different classifiers.

### 3.1 Data Collection & Pre-processing

We collected over 4 billion random English tweets posted after 2011 from Archive Team [16]. To train the model, we used the *text* of tweets and discarded the metadata.

Raw tweets are highly unstructured and contain redundant information. Therefore, we pre-processed the data by taking multiple steps as follows:

- **Removing unnecessary items:** We removed hashtag signs, reserved words (e.g. RT), HTML entities (e.g. <, >, &), punctuation, stopwords, and numbers, because they don't add any meaning to the tweets and keeping them may harm the embedding [17].

- **Text replacements:** It is common for people to mention other users' IDs in their tweets, but those user IDs should not be considered as a unique vocabulary in the embedding space. Therefore, we replaced mentions, URLs, and email addresses with "mentionn", "linkks", and "emailss", respectively.

- **Emoji separation:** Users usually do not consider emojis as separate words. Therefore, using emojis without any space between them and repeating the same emoji to emphasize a feeling is common in social media texts. In those cases, we added a space between emojis, and if a single emoji was repeated, we only kept one of them.

- **Removing redundant letters**: Users also change words with redundant characters to express their feelings. For example, instead of using the simple word "good", we find something similar to "goOoOoOoOoOoOod" in many tweets. Although finding all these redundant letters is challenging, we used a regular expression tool to uncover all three consecutive same characters and replace them with two consecutive ones.

For the cleaning process, we used a tokenizer developed by Erika Varis [18].

### 3.2 Modeling

After cleaning and tokenizing the data, we utilized the *word2vec* model from gensim to compute the word embedding from our data. We specified the hyperparameters as follows:

- **Embedding size:** We selected an embedding size of 300, similar to the *word2vec* model, and mapped *emojiSpace* to the Google News original pre-trained *word2vec* space [19] for easier use by other researchers.

- **Minimum count:** This hyperparameter specifies a threshold on frequency of word usage in the data. In the vocabulary list, the model excludes all the words that are repeated less than this threshold. Selecting small values for the minimum count, adds words used in very few tweets, and it may bias the embedding based on a small number of tweets. On the other hand, large values remove less frequent emojis and words. We iterated on the values for the minimum count and selected 50 as a reasonable value for this dataset.

- **Window size:** Some users prefer to use emojis as an immediate indicator of their feelings, and others prefer to add different emojis together in one place. We utilized the window size of 10 for our model, which is large enough to cover both cases.

By training the model using the above hyperparameters, we got an embedding with a vocabulary size of 2,011,787.

To justify our embedding, we use similarities of words. We found the most similar words to some emojis shown in Table 1. As we see in this table, the most similar words to the emojis have very identical themes. For example, 🤢 in the 5th row is close to "eww", "yuck", or "ew" that are excellent descriptions for this emoji.

---
[1]Gensim is a Python package that implemented *word2vec*, and it is available to the public.

|   | emoji | T1 | T2 | T3 | T4 | T5 | T6 | T7 | T8 |
|---|---|---|---|---|---|---|---|---|---|
| 1 | 😈 | 👹 | 👿 | 🤬 | 😠 | 🩸 | 😤 | 🔪 | 🔫 |
| 2 | 😭 | 😂 | 🙄 | 😩 | lmao | 🤣 | lmaoo | omg | 💀 |
| 3 | 💪 | 🙌 | ✊ | 🙌 | 🤟 | 🤙 | 🤘 | 👍 | 👌 |
| 4 | 😍 | 🥰 | 🤩 | 😻 | 💕 | 😏 | 💖 | 😊 | 😘 |
| 5 | 🤢 | 🤮 | eww | yuck | ew | eeww | eew | 😐 | nasty |
| 6 | 💒 | 👰 | 🕌 | 💍 | 🕎 | ⛪ | 👩 | 📿 | 🦋 |
| 7 | 😡 | 👿 | 😠 | 🙄 | 😑 | 😈 | 😤 | 👹 | 😒 |
| 8 | 👰 | 🦋 | 🧍 | 💒 | 💍 | 👭 | 👫 | 🧍 | cutestarn |
| 9 | 🛵 | 🛴 | 🚗 | 🚐 | 🚑 | 🚲 | 🚉 | 🚠 | 👟 |
| 10 | ❤️ | 💖 | 💕 | 💕 | 🥰 | ❤️ | 🤍 | 💝 | 😘 |
| 11 | 🤩 | 😍 | 😻 | 😎 | 🥰 | 🔥 | ✨ | 😻 | 🎉 |
| 12 | 🌸 | 🌷 | 🌸 | 🌼 | 🌹 | 🌻 | 🌿 | 🌱 | 💐 |

Table 1: Most similar emojis and words to a sample of 10 emojis.

| *word2vec* | | *emojiSpace* | |
|---|---|---|---|
| word | similarity | word | similarity |
| **we** | 0.65 | **them** | 0.64 |
| **them** | 0.64 | you | 0.58 |
| They | 0.62 | **themselves** | 0.57 |
| **their** | 0.61 | **we** | 0.56 |
| do | 0.57 | **their** | 0.55 |
| not | 0.56 | yall | 0.54 |
| theirs | 0.55 | ppl | 0.47 |
| have | 0.55 | em | 0.47 |
| **themselves** | 0.55 | people | 0.46 |
| ifthey | 0.55 | theyll | 0.45 |

Table 2: Evaluating our embedding based on similarity of the words to the word "they".

| Emoji | 1 | 2 | 3 | 4 | 5 | 6 | 7 | 8 | 9 | 10 |
|---|---|---|---|---|---|---|---|---|---|---|
| Common | 😄 | ❤️ | ❤️ | 💕 | 😂 | 😊 | 😍 | 😒 | 😭 | 🙈 |
| Rare | 🚟 | 🚞 | 🔤 | 🚝 | 🚡 | 🕰 | 🍅 | 🗾 | ⛎ | 🕐 |

Table 3: Examples of 10 common and rare emojis based on *emojitracker* [21]

Likewise, we can find the most similar words to commonly used words. For example, in Table 2 we are getting the most similar words to the word "they" and comparing our similarity results with *word2vec*. In this table, we highlighted the words that are shared in both embeddings. Checking the most similar words generated by *emojiSpace*, we can observe that the closest *informal* words represented here are similar to what we expect. Note that we are using Twitter data that has informal language compared to *word2vec* which is trained on Google News that has a formal language. Thus, language models build on *emojiSpace* can get better results in the sentiment analysis of informal conversations such as social media posts.

## 4 Evaluation

To evaluate the resulting embedding, we used sentiment analysis as an extrinsic task. For this task, we replicated the evaluation part of *emoji2vec* [15] and compared it with *emojiSpace*. For this downstream task, we used the dataset provided by Kralj Novak et al. [20] that consists of over 67 k English tweets labeled positive, negative, and neutral. In both training and test set, 29% of data is labeled as positive, 25% as negative, and 46% of tweets as neutral. Since the labels are almost evenly distributed (between positive and negative labels), accuracy is an effective metric in determining performance in this classification task.

We evaluated our model with and without emojis to find out how the performance of our model is improved by using emojis. We used two different classifiers, RF and SVM, to evaluate which one would outperform another. To compare the performance of our model with other models, we used two other sets of pre-trained embeddings as well. The first one is the original Google News *word2vec* embedding [19] and the second one is *word2vec* augmented with *emoji2vec* trained from Unicode descriptions [15]. In the training process, we defined the feature vectors by summing up the embedding vectors corresponding to each word or emoji in the tweet's text.

We used *emojitracker* website which ranks frequency of emojis usage over the web [21]. This website ranks a list of 845 emojis based on their frequency. We divided this list of emojis into two subsets: the top 20% of the list (173 most frequently used emojis) and the bottom 80% (672 less frequently used emojis). Hereinafter, the former group will be referred to as the "common" emojis, and the latter group will be referred to as the "rare" emojis. Based on these two groups of emojis, we defined two subsets. First, tweets containing common emojis and second, tweets containing rare emojis.

To find how much improvement we can get on sentiment classification by using emojis, we considered two scenarios. One uses the embedding of all the words in the tweet and the other uses the embedding of words and emojis in a tweet. We did the classification in these two scenarios and summerized the result in table 4. As we can see from the result in table 4, removing emojis had a negative effect on the task performance.

The *emojiSpace* outperforms *emoji2vec* in all the subsets we tested. Table 5 shows the results of using each embedding for sentiment analysis on the whole tweets and the two subsets. As we can see, the

|        | RF  | SVM | RF 😊 | SVM 😊 |
|--------|-----|-----|-------|--------|
| Common | 39% | 46% | 58%   | 63%    |
| Rare   | 42% | 44% | 54%   | 59%    |

Table 4: The effect of adding *emojiSpace* emoji-embedding on sentiment classification accuracy of RF and SVM in different scenarios. One subset contains tweets with common emojis, and the other contains the same tweets without emojis. The first two columns only use the embedding of the words, and the last two columns use *emojiSpace* emoji and word-embedding.

*emojiSpace* embedding outperforms two other embeddings. Also, our findings indicate that, in all scenarios, the linear SVM classifier outperforms the RF in this sentiment analysis task.

## 5 Conclusion and Future Work

Although users of many social media applications use emojis to express their feelings, most NLP projects in this domain still utilize word embeddings such as *word2vec* and *Glove* that do not present emoji embeddings. In this project, we trained a new embedding using a large number of tweets that include emojis. To the best of our knowledge, the number of tweets used in this project is more than any dataset used for emoji embeddings. *emojiSpace* contains embedding for both words and emojis and is obtained using the genism method on 4 billion English tweets containing emojis. For evaluating *emojiSpace*, we used sentiment analysis as the downstream task using two different classifiers of RF and SVM, and found the following results:

- We compared the performance of these two classifiers and found that Linear SVM outperformed RF classifier in all the scenarios that we used in the evaluation.

- We compared *emojiSpace* performance with two other pre-trained embeddings (Google News original *word2vec*, and *word2vec* augmented with *emoji2vec*) on sentiment analysis task. When we used all tweets and tweets with rare emojis as the test set, *emojiSpace* outperformed the two others pre-trained embeddings. Only in the scenario of using tweets with common emojis, Google News *word2vec* augmented with *emoji2vec*) outperformed *emojiSpace* if linear SVM was used as the classifier. Using RF as the classifier in this scenario, resulted in the better performance of *emojiSpace*.

- We compared the performance of *emojiSpace* on the same subsets of tweets with and without emojis, and the results showed that removing emojis from the tweets, decreases the performance of both of the classifiers in the sentiment analysis task.

Based on these results that met our expectation of improving the classifiers' performance in the sentiment analysis task, we validated *emojiSpace* reliability.

For future works, it is possible to use a "translation matrix" (or any other transformations) to map *emojiSpace* word-embedding into other embedding spaces. This helps to use emoji-embedding of *emojiSpace* together with those embeddings. As an alternative approach, it is possible to use the most similar words to emojis from *emojiSpace* to find emoji-embeddings similar to *emoji2vec* approach.

We believe there is still space to work more rigorously on pre-processing of the data used in this project. Specifically, for any NLP task on Twitter posts, it is essential to note that users are using a decent amount of slang and abbreviations. For example, "DIAF" stands for Die in a Fire, which clearly represents negative sentiment, or "HT", which stands for the hat tip, which indicates giving credit to another person and represents positive sentiment. Therefore, in future works, it is possible to collect the most frequent slang used in Twitter posts, replace them with their representative words, and see their impact on producing better word embedding. This project was not focused on the sentiment analysis part, so building on top of this embedding can result in much better sentiment analysis.

## Acknowledgment

The authors would like to thank Yangfeng Ji from the University of Virginia for his great suggestions about this course project.

|  | All tweets | | Common | | Rare | |
| --- | --- | --- | --- | --- | --- | --- |
| **Embeddings** | **RF** | **SVM** | **RF** | **SVM** | **RF** | **SVM** |
| Google News | 58% | 61% | 47% | 50% | 44% | 44% |
| Google News + *emoji2vec* | 59% | 62% | 50% | 63% | 47% | 49% |
| *emojiSpace* | 60% | 62% | 58% | 63% | 54% | 59% |

Table 5: Comparing *emojiSpace* sentiment classification accuracy with Google News and Google News + *emoji2vec* on the whole tweets dataset and the two subsets of tweets containing common emojis and tweets containing rare emojis.